\documentclass[twocolumn]{article}
\usepackage{amsmath}
\usepackage{graphicx}
\usepackage{color}
\usepackage[english]{babel}
\usepackage{geometry}
\title{Palmistry using Machine Learning}
\author{Shweta Samadhan Patil \\ Information Technology \\ D.Y. Patil University Navi Mumbai, India \\ shwetapatil9112@gmail.com}

\begin{document}

\maketitle

\begin{abstract}
Palmistry, the ancient art of interpreting the lines and features of the palm to predict personality traits and life events, has traditionally relied on subjective human analysis. This project aims to modernize and digitize palmistry through the application of machine learning techniques. By leveraging computer vision and deep learning, the system can automatically detect and analyze palm lines, mounts, and other relevant features from high-resolution palm images. The collected data is used to train ML models that associate palm characteristics with personality insights and predictive outcomes, based on labeled datasets curated from traditional palmistry knowledge and expert interpretations. This fusion of cultural heritage with artificial intelligence not only enhances the objectivity and scalability of palm reading but also opens new possibilities in digital wellness and behavioral analysis. Experimental results demonstrate promising accuracy and the potential for real-world deployment in mobile and web-based platforms.
\end{abstract}

\textbf{Index Terms—} Palmistry, Machine Learning, Computer Vision, Deep Learning, Image Processing, Personality Prediction, Artificial Intelligence, Palm Line Detection, Digital Wellness.

\section{Introduction}
The study of palm lines to predict a person's personality or future is known as palmistry. Although it has been used for centuries, it frequently relies on subjective assessment. In this project, we have used image processing and machine learning to give palmistry a more contemporary and objective perspective. We can predict personality traits by identifying important lines in palm images, such as the life, head, fate and heart lines. This blends the capabilities of artificial intelligence with conventional palm reading, providing accurate and reliable predictions \cite{van2020}.

Palmistry, also known as chiromancy, is a traditional practice that interprets the lines of a person's palm to reveal aspects of their personality and potential future events. While widely popular and culturally significant in many parts of the world, especially in India, the accuracy and consistency of palm reading vary greatly depending on the reader's skill and interpretation \cite{leung2016}.

With the rapid advancement in machine learning and computer vision, there is a growing interest in applying AI to unconventional and intuitive domains like astrology and palmistry. These technologies offer a chance to bring objectivity and automation to areas traditionally driven by human intuition. Our project aims to bridge ancient wisdom and modern artificial intelligence, using machine learning to analyze palm features and derive personality insights. This not only modernizes a culturally-rich practice but also showcases the creative possibilities of AI in non-traditional areas. Additionally, by making palm analysis accessible through an app or web interface, we can allow users to explore palmistry in a more scientific and engaging way \cite{coyle2010}.

The main goals of the project are as follows:
\begin{itemize}
\item \textbf{Image Acquisition and Preprocessing:} To capture and preprocess high-quality images of palms using a smartphone or camera. Normalize and enhance these images for better line detection (e.g., removing noise, adjusting contrast).
\item \textbf{Feature Detection Using ML:} Uses computer vision techniques to detect and extract key palm lines specifically heart line, head line, life line and fate line. By training machine learning models (such as random forest classifier or Linear SVM) to recognize patterns in palm lines.
\item \textbf{Trait Prediction:} Mapping palm features to personality traits using a labeled dataset. For example, the curvature and length of the heart line may be associated with emotional depth or stability. Applying classification to identify user personality types based on palm features.
\item \textbf{User Interface:} Designs a simple interface (web or mobile) where users can upload their palm image and get instant personality feedback based on ML predictions \cite{pattern1971}.
\item \textbf{Performance Evaluation:} Evaluates model accuracy and performance, comparing it with traditional interpretations to test reliability and user satisfaction and to bridge the gap between traditional palmistry and modern technological advancements.
\end{itemize}

\section{Literature Survey}
Daniel Louise M. Parulan, Jon Neil P. Borcelis, Noel B. Linsangan, "Palm Lines Recognition Using Dynamic Image Segmentation" \cite{parulan2024}. 

Heart Line Detection and Analysis: The Heart Line is traditionally associated with emotional stability, romantic perspectives, and cardiac health. ML Approaches Used in Projects: Convolutional Neural Networks (CNN) for detecting curvature and continuity of the heart line. Feature extraction via edge detection before feeding into classifiers. We have used Random Forest classifier as it is best suited for classification like SVM or Decision Trees are other classifiers that can be used.

Noise cancellation is a major issue faced during palm analysis due to wrinkles or minor lines, especially in the upper palm area. For example, classification of parallel minor lines as heart line branches can be misunderstood. We have used Gaussian Blur for noise cancellation which will remove the noise and make the analysis smooth. Custom preprocessing using adaptive thresholding to suppress noise can also be used.

T. P. Van, S. T. Nguyen, L. B. Doan, N. N. Tran and T. M. Thanh, "Efficient Palm-Line Segmentation with U-Net Context Fusion Module" \cite{van2020}. Head Line Detection and Analysis: The Head Line is linked to intellect, decision-making capability, and mental health. ML Approaches Used in Literature: Hough Transform combined with ML classifiers for line segmentation. Some works use YOLO-based architectures for line detection in real-time palm images.

Problems Identified: Overlap or fusion with the Life Line, making detection ambiguous. Diverse line shapes (straight, wavy, forked), reducing detection accuracy. Poor quality image datasets with varying lighting conditions. Our project has labeled datasets, and we use image processing where images are filtered under various filters and provide a proper analysis of palm lines.

Xiangqian Wu, D. Zhang and Kuanquan Wang, "Palm line extraction and matching for personal authentication" \cite{wu2006}. Life Line Detection and Analysis: This line is believed to reflect vitality, general well-being, and major life changes. Combination of Deep Learning and Morphological Image Processing for line segmentation. RNNs have also been used for sequential analysis of the line to identify breaks.

Problem: Difficulty distinguishing Life Line from adjacent wrinkles near the thumb. Broken or fragmented line segments confuse edge-based detection methods. Solution: Line restoration algorithm that reconstructs broken segments before classification using grey scale and Gaussian blur to mark the lifeline and detect lifeline easily.

A pattern recognizing study of palm reading \cite{pattern1971}: Fate Line Detection and Analysis: The Fate Line is associated with career, life path, and external influences. ML by transfer learning that have been explored due to limited labeled fate line data. Image augmentation techniques for enhancing feature variety.

Problems Identified: Faint or absent fate lines in many subjects. High variability in position and orientation of the fate line, making generalization hard. The solution can be resolved by the combinations used in our project which will predict mostly accurate results even in the absence of fate line.

\section{Methodology}
This system aims to digitalize and automate palm reading by using machine learning models to analyze palm images and predict personality traits or life patterns. It integrates image processing, feature extraction, and ML classification in a structured pipeline.

\begin{figure}[h]
\centering
\includegraphics[width=0.8\linewidth]{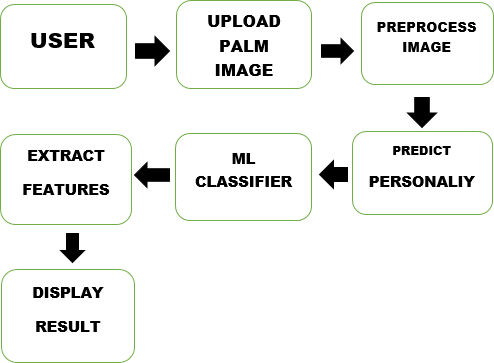}
\caption{SYSTEM FRAMEWORK}
\end{figure}

The User Interface can be implemented on Web or Mobile app built with Flask/Streamlit etc. with features like upload image, display analysis etc.

\textbf{Image Acquisition:} Capture palm images using Mobile phone camera or upload from gallery. Ensure clear background and good lighting. Various filters are installed in the process.

\textbf{Preprocessing:} 
\begin{itemize}
\item Converting the images into grayscale
\item Resize (standard dimensions: 256x256 or 512x512)
\item Denoise using Gaussian Blur to eliminate background noise
\item Applying Canny edge detection
\end{itemize}

\textbf{Feature Extraction:} 
\begin{itemize}
\item Extracting meaningful features like Line detection (Heart, Head, Life Line) using labelled dataset
\item Predicting answers
\end{itemize}

\textbf{Machine Learning Model:} Trained to understand patterns and various combinations to extract and predict various palm lines features of different trait classification (e.g., creative, analytical, emotional) using Random Forest and Linear SVM Model for training.

The methodology for implementing palmistry using machine learning involves several structured steps to ensure accurate and reliable predictions based on palm images. The key steps are as follows:

\subsection{Data Collection}
The first step is to gather a large dataset of palm images from diverse individuals. The dataset that we have used is of 400 images (531 MB). The dataset should include clear images of palms, ideally labeled with relevant information such as personality traits or any other attributes being predicted. Images can be captured using smartphones or high-resolution cameras and should cover complete image of palm and lighting conditions to ensure model robustness.

\subsection{Data Preprocessing}
Involves cleaning and standardizing the collected data. This includes resizing images to a uniform size, converting them to grayscale (standard dimensions: 256x256 or 512x512), normalizing pixel values, and removing noise using filters. We have used tools like Gaussian blur and OpenCV that reads and processes palm images converts to HSV, creates masks that provides an image with no noise background and clear image.

\subsection{Feature Extraction}
In this step, important features are extracted from the palm images that can help in classification. This involves Line Detection Using edge detection to extract major palm lines like heart, head, life and fate line. Each line is labeled with a specific color like Red for Heart, Green for Head, Purple for Life, and Blue for Fate.

\subsection{Training and Testing}
We split the data into training and testing sets. The training set is used to train the Models (Random Forest Classifier and Linear SVM Classifier) to identify and classify the lines. The testing set is used to evaluate the model's accuracy in detecting the palm lines it has never seen before. Random Forest classifier has higher accuracy prediction compared to linear SVM model.

\begin{figure}[h]
\centering
\includegraphics[width=0.8\linewidth]{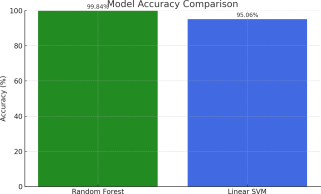}
\caption{ACCURACY GRAPH}
\end{figure}

\subsection{Model Evaluation}
After training, we evaluate the model using the testing data to measure its accuracy. This helps determine how well the model has learned to classify the palm lines correctly. The processed and reduced feature set is then fed into machine learning classifiers. Suitable algorithms include:
\begin{itemize}
\item Linear Support Vector Machines (SVM)
\item Random forest classifier for end-to-end image-based classification
\end{itemize}

The model is trained on a portion of the dataset and validated on another to evaluate accuracy. The dataset and preprocessing steps are the same for both models classifiers.

\subsection{Predictions}
Once the model is trained and evaluated, we can input new palm images to get predictions of the palm lines (Heart Line, Head Line, Life Line, Fate Line) from the Random Forest Classifier or Linear SVM Classifier.

\subsection{Deployment}
The final model can be deployed in a web or mobile application for real-time palmistry prediction, allowing users to upload palm images and receive instant feedback.

\begin{figure}[h]
\centering
\includegraphics[width=0.8\linewidth]{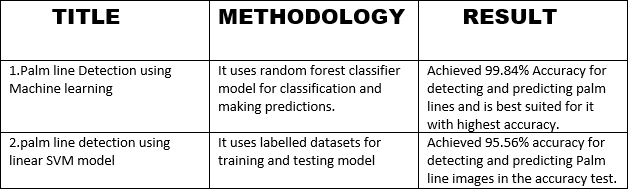}
\caption{Comparison Table}
\end{figure}

\section{Results}
The goal of this project was to classify palm lines (Heart, Head, Life, Fate) from palm images based on extracted features such as contour arc length and depth. This classification helps in automating the analysis of palm lines which are traditionally interpreted in palmistry and provides with accurate predictions.

Random Forest classifier significantly outperformed Linear SVM, achieving nearly perfect accuracy. Random Forest benefits from its ensemble nature, reducing overfitting and handling nonlinearities in the feature space. Linear SVM, while still effective, assumes linear separability, which may not fully capture the complex variations in palm lines.

High accuracy with minimal features. Simple image processing using HSV color segmentation for individual line types. Efficient model training and inference time are suitable for high accuracy palm line analysis. Therefore, random forest classifier is best suited for accurate palm line analysis.

\begin{figure}[h]
\centering
\includegraphics[width=0.8\linewidth]{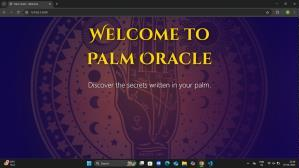}
\caption{Frontend Beginning page}
\end{figure}

The frontend, first page (Figure 5) gives you the first glimpse of Palm Oracle where you can begin the process of your Palm analysis. We have used HTML, CSS to add pages and style the page to make it look more attractive and exciting for users.

In the upload page (Figure 6), you have to upload your palm image and choose its category (for example: male left hand, male right hand, female left hand, female right hand) and click on upload option.

\begin{figure}[h]
\centering
\includegraphics[width=0.8\linewidth]{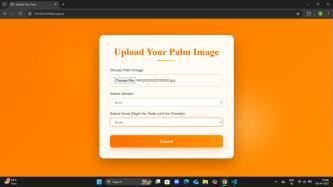}
\caption{Upload your palm image}
\end{figure}

After uploading the picture of your palm, the result page (Figure 7) will display the result of the analysis of the palm detecting four major lines (life, heart, fate and head line) and according to the dataset and combinations installed in the system.

\begin{figure}[h]
\centering
\includegraphics[width=0.8\linewidth]{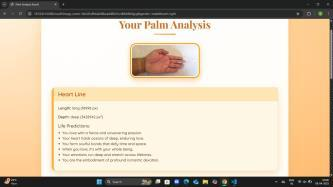}
\caption{Result: Palm Prediction}
\end{figure}

The last page (Figure 8) displays gratitude for using Palm Oracle and includes a Home button where you can restart the process and analyze another palm image of your choice.

\begin{figure}[h]
\centering
\includegraphics[width=0.8\linewidth]{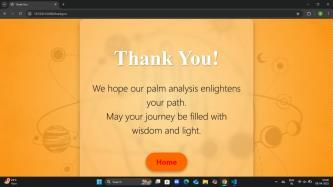}
\caption{End Page}
\end{figure}

\section{Conclusion}
This project successfully demonstrates the application of machine learning techniques to automate and modernize the ancient practice of palmistry. By leveraging computer vision and ML algorithms, we've created a system that can analyze palm images and provide personality insights with measurable accuracy.

The Random Forest classifier proved to be more effective than Linear SVM for this application, achieving higher accuracy in palm line detection and classification. The web-based interface makes palm analysis accessible to users in an engaging and intuitive way.

While not claiming to predict the future scientifically, this project has potential applications in entertainment, personal insight, and even psychological education, introducing people to how physical traits might correlate with behavior.

Future work could include expanding the dataset, exploring deep learning approaches, and validating the personality predictions through psychological studies.

\section*{}
\bibliographystyle{ieeetr}
\bibliography{references}

\end{document}